\documentclass[runningheads]{llncs}

\usepackage{eccv}

\usepackage{comment}
\usepackage{hyperref}
\usepackage{eccvabbrv}
\usepackage{xcolor}

\usepackage{graphicx}
\usepackage{booktabs}
\usepackage{multirow}
\usepackage{overpic}

\usepackage[normalem]{ulem}
\useunder{\uline}{\ul}{}

\usepackage[accsupp]{axessibility}  

\makeatletter
\renewcommand\paragraph{\@startsection{paragraph}{4}{\z@}%
    {.5em}%
    {-1em}%
    {\normalfont\normalsize\bfseries}}
\makeatother

\newcommand{\compress}{\vspace{-3.15mm}}
\newcommand{\smallcompress}{\vspace{-2.1mm}}

\usepackage{hyperref}

\usepackage{orcidlink}

\usepackage{algorithm, algpseudocode}

\usepackage{booktabs} 
\usepackage{array}    

\begin{document}

\title{SAM4MLLM: Enhance Multi-Modal Large Language Model for Referring Expression Segmentation} 

\titlerunning{SAM4MLLM}

\author{Yi-Chia Chen\inst{1} \and Wei-Hua Li\inst{1} \and 
Cheng Sun\inst{2} \and \\ Yu-Chiang Frank Wang \inst{1,2} \and Chu-Song Chen\inst{1}}

\authorrunning{Chen et al.}

\institute{National Taiwan University, Taipei, Taiwan. \\
\email{\{r11922a04, d12922009, chusong\}@csie.ntu.edu.tw}\\ \and
NVIDIA \\
\email{\{chengs, frankwang\}@nvidia.com}
}


\maketitle

\begin{abstract}

  We introduce SAM4MLLM, an innovative approach which integrates the Segment Anything Model (SAM) with Multi-Modal Large Language Models (MLLMs) for pixel-aware tasks. Our method enables MLLMs to learn pixel-level   location information without requiring excessive modifications to the existing model architecture or adding specialized tokens. We introduce an inquiry-based approach that can effectively find prompt points for SAM to perform segmentation based on MLLM. It combines detailed visual information with the powerful expressive capabilities of large language models in a unified language-based manner without additional computational overhead in learning. Experimental results on pubic benchmarks demonstrate the effectiveness of our approach.
  {Our code is available on \href{https://github.com/AI-Application-and-Integration-Lab/SAM4MLLM}{GitHub.}}
  
  \keywords{LLM \and MLLM \and Referring Expression Segmentation.}
  
\end{abstract}

\section{Introduction}
\label{sec:intro}


With the rapid growth of generative AI, large language models (LLM) \cite{floridi2020gpt, chowdhery2023palm, zhang2022opt, workshop2022bloom, touvron2023llama, touvron2023llama2, bai2023qwen_txt} become a focus of research and application due to their profound capabilities in understanding and generating text. They show innovative power in machine learning and marks the evolution of human-machine interaction. 


Recently, the progress has been made from simple text processing to the complex multi-modal understanding. 
The advent of Multi-modal Large Language Models (MLLMs) \cite{li2023blip, huang2024language, achiam2023gpt, liu2024visual, zhang2023llama, zhu2023minigpt, team2023gemini} lies in incorporating image processing modules into LLMs.
They successfully endow LLMs with the ability to process visual information, thereby bridging the significant gap between visual and linguistic tasks.
Recent studies enabled MLLMs to engage in in-depth dialogues based on image content.
Subsequent research enhanced MLLMs capabilities through data or structural modifications, leading to enhanced MLLMs that allow for the input and output of bounding boxes of objects to achieve fine-grained visual dialogues \cite{peng2023kosmos, chen2023shikra, zhang2023gpt4roi, bai2023qwen, lin2023sphinx, chen2023minigpt}.



{\color{black}Referring Expression Segmentation (RES)~\cite{wang2022cris, xu2023bridging, yang2022lavt, zou2024segment}} aims to label image pixels corresponding to specific objects or reuns mentioned in natural language expressions. It involves accurately identifying and segmenting the objects referred to by linguistic descriptions.
In this paper, we focus on RES and use MLLM to solve this task.
However, bounding boxes alone are insufficient for precisely indicating object locations within images. 
This has led to research efforts focused on improving annotation granularity to pixel-level for MLLM, enhancing image information encoding, and designing models capable of outputting detailed segmentation masks \cite{lai2023lisa, pi2023perceptiongpt, rasheed2023glamm, xia2023gsva}. 
Despite significant progress, these advancements 
require substantial modifications to the original MLLMs architectures \cite{rasheed2023glamm}. 
Some study 
introduces 
additional model structures to output segmentation masks \cite{pi2023perceptiongpt}; 
some others leverage the use of special tokens different from those in original LLMs \cite{lai2023lisa, xia2023gsva} or 
rely on the application of multiple losses 
for model optimization \cite{lai2023lisa}. 
These adjustments introduce architectural complexity to MLLM and may complicate model extension for additional tasks.

In this paper, we propose a {\color{black}simple solution} that can enhance MLLM's abilities to understand object localization in pixel level.
Our approach is simple but effective, 
which upgrades MLLMs' visual 
capabilities to a new level for accurate understanding the referring expressions of pixel-level location 
in images. 

Our method draws inspiration from the context provided below. Concurrently with the development of LLMs and MLLMs, the field of image segmentation has witnessed a significant breakthrough with the introduction of the Segment Anything Model (SAM) \cite{kirillov2023segment}, a foundation model trained on the SA-1B\cite{kirillov2023segment} high-quality image segmentation dataset. SAM, a promptable segmentation model, can generate high-quality semantic-free segmentation masks in images based on prompts provided by the user, such as points or bounding boxes.

We observe that while MLLMs possess a profound understanding of image semantics, they struggle to articulate detailed pixel-level information. Conversely, SAM, although not semantically aware, can delineate intricate segmentation masks with minimal prompting. In light of this, we propose a novel methodology using SAM for MLLM (namely, SAM4MLLM) which seamlessly integrates MLLMs with SAM. Specifically, we employ a straightforward yet {\color{black}simple strategy}, introducing pixel-level information into the training dataset without altering the original MLLM architecture. This enables MLLMs to grasp pixel-level information using the same text cross-entropy loss used by {\color{black}popular LLMs~\cite{floridi2020gpt, jiang2023mistral, bai2023qwen, touvron2023llama2}}. 
Considering the potential limitations of MLLMs in pixel expression due to input resolution constraints and a model architecture not explicitly designed for visual tasks, we further enhance the output with SAM, post-processing MLLM outputs to obtain higher precision segmentation masks in a relatively effortless manner. To establish a connection between SAM and MLLM, one straightforward approach is to enable MLLM to generate prompt points for SAM. However, effectively producing multiple points can be challenging. Therefore, we introduce a novel method that leverages the dialog capability of LLMs. Specifically, we proactively ask the MLLM to acquire effective prompt points for SAM.
We tackle the problem of RES and demonstrate the effectiveness of our approach.

Main contributions of this work are as follows:

\noindent $\bullet$~~ We present SAM4MLLM, an approach allowing MLLMs to understand pixel-level details without altering the MLLM model architecture, introducing new tokens, or employing additional losses. It is simple yet effective for RES.

\noindent $\bullet$~~~To connect MLLM and SAM, we introduce a novel method of actively querying the language system to obtain prompt point cues.

\noindent $\bullet$~~ Through experiments on various RES benchmarks, including RES dataset, GRES, and ReasonSeg, we validate the effectiveness of SAM4MLLM and demonstrate its favorable performance in handling complex pixel-aware tasks.


\section{Related Works}
In this section, we review the related works of the topics: RES, image segmentation, MLLMs, and MLLMs toward segmentation.

\paragraph{Referring Expression Segmentation.}
Early researches in RES focus on integrating features from language and vision models to effectively merge these two types of information. 
Yu~\emph{et al.}~\cite{yu2018mattnet} combine the language attention and subject, location, and relationship modules to localize the target region. 
In STEP~\cite{chen2019see}, a DNN architecture is applied to iteratively refine the segmentation heatmap. 
Subsequently, Hui~\emph{et al.}~\cite{hui2020linguistic} introduce Linguistic Structure guided Context Modeling (LSCM) to aggregate the multi-modal features. To understanding the language expression in different perspective, VLT~\cite{ding2021vision} generates several sets of queries and introduces a query balance module to focus on the most suitable query.
Zhu~\emph{et al.}~\cite{zhu2022seqtr} regard RES as a point prediction problem and design a simple transformer-based network to perform referring segmentation. 
Recently, the method in \cite{xu2023bridging} leverages a novel adapter to facilitate cross-modal information. These studies lay the foundation for subsequent work on MLLMs in RES.

With the advancement of multi-modal model, they have been introduced into the RES field, enhancing the accuracy and efficiency of segmentation.
Wang~\emph{et al.}~\cite{wang2022cris} introduce the multi-modal model CLIP~\cite{radford2021learning} to RES tasks. 
With the rise of MLLMs, research based on these models has emerged, leveraging their remarkable abilities in understanding text and images.~\cite{lai2023lisa, xia2023gsva, pi2023perceptiongpt, rasheed2023glamm}. 
Furthermore, it has been pointed out in \cite{liu2023gres} that classical RES benchmarks are not sufficiently comprehensive in some cases, leading to the proposal of General Referring Expression Segmentation (GRES) \cite{liu2023gres} dataset to broaden its application scope. 
GRES allows for the reference to multiple objects simultaneously and can address the absence of objects in images, further enhancing the 
applicability in practice.
In LISA~\cite{lai2023lisa}, a more complex dataset, ReasonSeg, is proposed. It requires models to possess complex reasoning abilities and a basic understanding of the real world, addressing challenges that are closer to real-world scenarios. 

\paragraph{Image Segmentation and Segment Anything.}

Image segmentation is a central task in computer vision, aiming to identify and label objects within images at the pixel level. 
Methods like Fully Convolutional Networks~\cite{long2015fully}, Mask R-CNN~\cite{he2017mask} and Masf2Former \cite{cheng2022masked} have greatly advanced the field.
Recently, the Segment Anything Model (SAM)~\cite{kirillov2023segment} is trained on the SA-1B~\cite{kirillov2023segment} dataset with one billion high-quality segmentation annotations. SAM can segment high-quality object masks based on simple prompts. EfficientViT-SAM~\cite{cai2022efficientvit} further introduces multi-scale linear attention into the ViT backbone of SAM, increasing the speed of SAM by several times without compromising performance. Our SAM4MLLM employ 
MLLMs to guide SAM 
for precise object segmentation.

\paragraph{Multimodal Large Language Models (MLLMs).}
LLMs have proven their exceptional capabilities in the domains of language understanding and generation, with notable examples including GPT-3 \cite{floridi2020gpt}, BLOOM~\cite{workshop2022bloom}, PaLM~\cite{chowdhery2023palm}, OPT~\cite{zhang2022opt}, LLaMA~\cite{touvron2023llama}, LLaMA-2~\cite{touvron2023llama2}, Mistral~\cite{jiang2023mistral}, Qwen~\cite{bai2023qwen}, and others, significantly advancing the field of natural language processing. 
These models have not only demonstrated near-human levels of proficiency but have also spurred interest in the study of visual-language interaction, leading to the development of MLLMs. MLLMs are built upon 
LLMs by integrating innovative techniques that combine visual and linguistic modalities, such as the Perceiver Resampler introduced by Flamingo~\cite{alayrac2022flamingo}, the prompt tuning token by LLaMA-Adapter~\cite{zhang2023llama}, the Q-Former by BLIP-2~\cite{li2023blip}, and the use of linear projection layers in LLaVA~\cite{liu2024visual} to enable LLMs to interpret images. 

\paragraph{MLLMs Toward Segmentation.}
In 
MLLMs, researchers 
focus not only on enhancing the model's understanding of multimodal data but 
also empowering MLLMs with the capability to process detailed information. 
For instance, DetGPT~\cite{pi2023detgpt} introduces a method that combines MLLMs with open-vocabulary object detectors. GPT4RoI~\cite{zhang2023gpt4roi} incorporates region-of-interest information into instructions. Kosmos-2~\cite{peng2023kosmos} constructs a large-scale grounding image-text pairs dataset, named GRIT, which assists MLLMs in understanding regional information within images. Shikra~\cite{chen2023shikra} encodes all regional information in a linguistic form, eliminating the need for introducing new vocabulary, position encoders, or decoders to MLLMs. Ferret~\cite{you2023ferret} uses a hybrid regional representation method that combines discrete coordinates with continuous features to describe regions within images. However, the model outputs of these methods are limited to bounding boxes and have not yet achieved pixel-level precision operations.

Building on this foundation, Lai~\emph{et al.}~\cite{lai2023lisa} 
propose a method based on introducing \texttt{[SEG]} tokens and a SAM decoder, enabling MLLMs to perform reasoning segmentation tasks. 
In PerceptionGPT~\cite{pi2023perceptiongpt}, a lightweight visual task encoder and decoder are adopted to handle segmentation masks, allowing MLLMs to input and output segmentation masks. 
Ren~\emph{et al.} 
encode masks within images into segmentation codebooks, coupled with a lightweight decoder for mask output. 
GSVA~\cite{xia2023gsva} extends upon LISA~\cite{lai2023lisa} by supporting multiple \texttt{[SEG]} tokens and introducing \texttt{[REJ]} tokens, applying MLLMs to GRES tasks. 
Rasheed~\emph{et al.}~\cite{rasheed2023glamm}  
propose a Grounding LMM model capable of generating natural language responses seamlessly integrated with corresponding object segmentation masks.

Although the aforementioned models can output masks, they require modifications to the original MLLM architecture or the addition of new model structures to output masks, or the introduction of special tokens 
not belonging to the original LLMs. 
They may need to utilize multiple loss functions for simultaneous model optimization, 
which increases the complexity of MLLM design and 
poses obstacles to the model's expansion to more tasks. 
Our SAM4MLLM does not 
have these burdens, merely integrating with the off-the-shelf SAM model to output high-quality segmentation masks, thereby providing a new solution path for complex pixel-level tasks.

\section{Method}


In this section, we present our SAM4MLLM method. We first introduce how to encode segmentation masks with SAM's prompt, and then our solutions for prompting SAM using MLLM.

\subsection{Encode Segmentation Mask into SAM Prompt}

Existing MLLMs for segmentation (e.g., LISA \cite{lai2023lisa}, PerceptionGPT \cite{pi2023perceptiongpt}, GLaMM \cite{rasheed2023glamm}, GSVA \cite{xia2023gsva}) 
rely on specialized design of model architectures, 
segmentation-specific tokens, and 
heterogeneous loss functions 
to predict object masks. 
E.g., LISA~\cite{lai2023lisa} introduces a special token \texttt{[SEG]} and the associated architecture. 
It uses dice 
and binary-cross-entropy losses for segmentation, combined with text 
loss for training. 
This increases the model complexity and optimization difficulty. 

Our method leverages SAM's characteristic that it can convert few discrete text prompt tokens (\ie, bounding box plus several points indicating whether they are inside or outside the object region) to high-quality continuous-boundary segmentation masks.
Our \textbf{SAM4MLLM} uses the discretized image coordinate for points.
We encode an arbitrary-shaped mask by using a bounding box and \(\mathcal{K}\) points. The bounding box is expressed as \(Prompt_B \in \mathbb{N}^4\); the prompt of \(\mathcal{K}\) points, each of which contains three values, \(x\) coordinate, \(y\) coordinate, and whether the point is on the object mask, are encoded as \(Prompt_P \in \mathbb{N}^{\mathcal{K} \times 3}\).

By encoding continuous segmentation masks into discrete SAM prompts, 
we avoid adding any tokens or altering the model structure, while maintaining training with only text auto-regression cross-entropy loss.  
This method is 
consistent with the original training mode of language models, enabling MLLMs to understand pixel-level information and facilitate easier future model expansion.

\begin{figure}[t]
    \centering
    \includegraphics[width=1\linewidth]{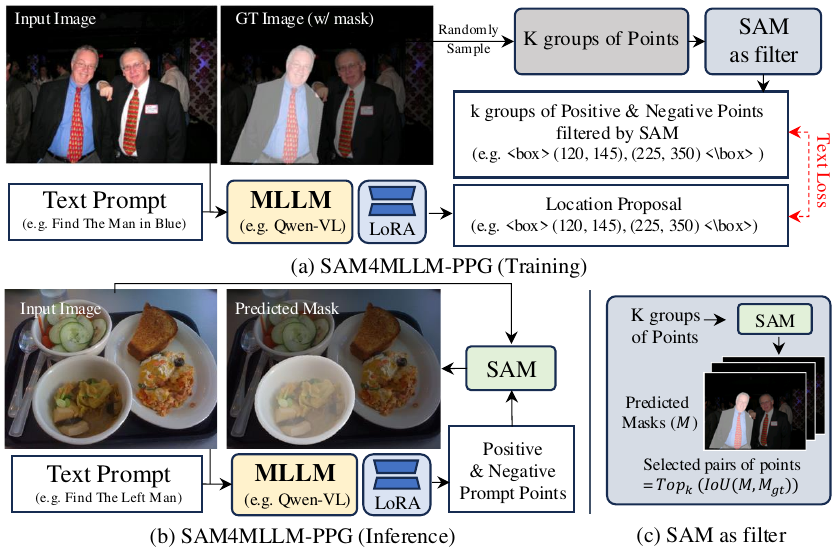}
    \smallcompress
    \caption{Architecture diagram of SAM4MLLM-PPG. 
    (a) The training process of PPG, 
    (b) The inference process of PPG, (c) SAM as filter.
    }
    \label{fig:SAM4LLM-PPG}
    \compress
\end{figure}

\subsection{Prompting SAM Using MLLM}


To incorporate SAM into MLLM in a unified way, a main issue lies in acquiring the prompt points for SAM, including the points that are positive (inside) and negative (outside) the object mask region. To do this, we introduce two solutions, \textit{Prompt-Point Generation} (PPG) and \textit{Proactive Query of Prompt-Points} (PQPP).
The former directly generates the proposal points by using the MLLM model in the inference stage. The later, on the other hand, acquires the points in an indirect manner; it uniformly samples the points in the bounding box at first, and then for each point asks the MLLM model whether the point is inside the object region or not. We respectively introduce them in the following.

\paragraph{SAM4MLLM-PPG.} In this method, an MLLM that can take both text-prompt and image inputs is adopted. To align the MLLM with segmentation utility, we use the parameter-efficient fine tuning technique, LoRA \cite{hu2021lora}, to train the model based on some RES datasets with image-text pairs and ground-truth masks. LoRA outputs the location prompt including the bounding box \(Prompt_B \in \mathbb{N}^4\) and \(k\) groups of positive and negative points \(Prompt_P \in \mathbb{N}^{(n_1+n_2)k \times 3}\), as illustrated in \cref{fig:SAM4LLM-PPG}(a), where a group contains $n_1$ positive and $n_2$ negative points ($n_1=2, n_2=1$ in our implementation). 

To provide the location supervision to LoRA, we randomly sample $K$ groups of points ($K>k$) in the training stage based on the object mask and then send these prompts to SAM. For every group, SAM delivers the segmentation result. 
We filter out the prompts with low IoUs compared to the ground-truth masks and only keep the top-$k$ groups (\cref{fig:SAM4LLM-PPG}(c)). 
In our implementation, only text loss (auto-regression cross-entropy loss) is required; $K$ is typically 64 and $k=1$. 
In the inference stage, LoRA directly delivers the points that are sent to SAM for segmentation, as shown in \cref{fig:SAM4LLM-PPG}(b). 
More details an be found in Sec.~\ref{exp:implementation_details}.

\begin{figure}[t]
    \centering
    \includegraphics[width=1\linewidth]{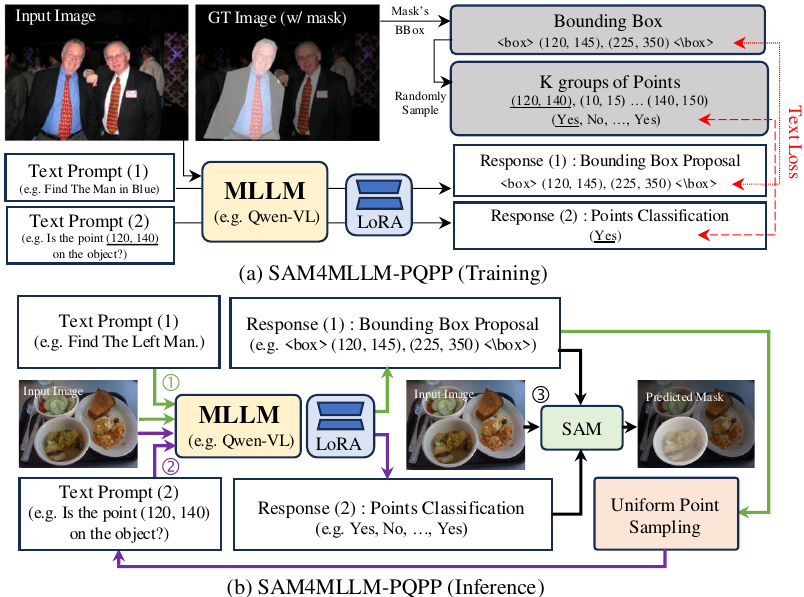}
    \smallcompress
    \caption{{\color{black}Architecture diagram of SAM4MLLM-PQPP. 
    (a) The training process of PQPP,
    (b) The inference process of PQPP.}
    }
    \label{fig:SAM4LLM-PQPP}
    \compress
    \vspace{-1.2em}
\end{figure}

\paragraph{SAM4MLLM-PQPP.} In this method, instead of producing the prompts directly, we propose to leverage the power of MLLM's query-response capability. We sample the prompt points and proactively ask the MLLM if they are inside (or outside) the mask. In the training phase, a bounding box and $K$ groups of points are randomly sampled based on the ground-truth mask, and a dialog containing two rounds are conducted. In the first round of the dialog, LoRA responses a bounding box. In the second round, for each of the $(n_1+n_2)K$ points, LoRA responses whether it is inside the mask (yes or no) during training.

In the inference phase, LoRA outputs a bounding box in the first round for the input text query and image. We then uniformly sample the points in the bounding box. 
The sampled grid points are sent to MLLM-LoRA again in the second round and asked if they are positive (or negative); the answers are applied to SAM for segmentation. 
We typically set the grid size as $5\times 5$. 
The training and inference phases of SAM4LLM-PQPP are illustrated in \cref{fig:SAM4LLM-PQPP}. 
To provide high-quality prompt points before sending to SAM, the low-confident points will be removed.
More details of SAM4LLM-PQPP an be found in Sec.~\ref{exp:implementation_details}.

Characteristics of the two solutions are as follows. The PPG 
adopts directly the MLLM to produce 
prompt points 
in addition to the bounding box. 
Yet it would be challenging to learn for simultaneously producing multiple points. Hence, only few prompt points are used in PPG.  
The PQPP 
leverages MLLM's dialog capability. 
It can first inquiry a rough 
bounding box and then probe multiple points of interest in the bounding box through query-answering for prompting the SAM. 
We compare their performance in the experiments.
\vspace{-1em}

\subsection{RES Training}
\label{method:implementatoin_overview}

To align the foundational MLLM to the RES task, we use the datasets containing the RES-relevant examples to guide the model toward the goal. 
Three datasets are used for training our SAM4MLLM to align with the RES task. 
Two of them (RES dataset and gRefCOCO dataset) contain RES data with ground-truth masks. The third (VQA) is a visual dialog dataset without masks, employed to enhance further the general capability of joint vision-language understanding.
During training, to preserve the generalization ability of MLLM on images, 
we freeze most of the network parameters, 
and adjust only the visual resampler of MLLM together with the LoRA adapter. 
The datasets are briefed below.

\paragraph{Referring Expression Segmentation Datasets (RES dataset):} Each sample 
in this dataset provides an image accompanied by a phrase denoting a specific object in the image. 
The phrase corresponds to 
only one object. This 
dataset includes publicly available 
subsets, refCOCO~\cite{yu2016modeling}, refCOCO+~\cite{yu2016modeling}, and refCOCOg~\cite{mao2016generation}.
They are based on images from the MSCOCO~\cite{lin2014microsoft} but compiled through different annotation processes. 
The primary difference between RefCOCO+ and RefCOCO is that the former prohibits the use of location-based descriptions (e.g., "the person on the right side of the picture"), thereby compelling annotators to focus on describing the appearance features of objects. 
RefCOCOg provides longer and more detailed descriptions that cover not only appearance information but may also include actions, locations, and details about relationships with other objects.

\paragraph{Generalized Referring Expression Segmentation (GRES) Dataset \cite{liu2023gres}:} Similar to the RES dataset, each sample offers an image and a phrase describing the objects to segment. 
The difference is that the phrase may not be present in the image or may refer to multiple objects simultaneously. We use the publicly available gRefCOCO~\cite{liu2023gres} dataset for this task.
Our SAM4MLLM can naturally generate additional SAM prompts for segmenting multiple instances. In case where the queried objects are not present, we train our model to predict ``object not in the image.''

\paragraph{Visual Question Answering (VQA):} To maintain the visual dialogue capability of MLLM, we incorporated VQA data, specifically using the VQAv2\cite{antol2015vqa}. 

For all the datasets mentioned above, we 
do not use data augmentation during training because flipping and/or cropping may change the relative position or relationship of objects in the image.






\section{Experiments}

In this section, we outline the experimental setups for our SAM4MLLM method, covering the network architecture, implementation details, evaluation datasets, and analysis of experimental results.
\vspace{-1.2em}
\subsection{Implementation Details}
\label{exp:implementation_details}
\vspace{-5pt}
\paragraph{Network Architecture:}
We use Qwen-VL-7B-Chat~\cite{bai2023qwen} as our 
MLLM backbone architecture for its ability to output bounding boxes from the pre-training phase. Specifically, the LoRA adapter is configured as follows: LoRA rank is set to 256, LoRA alpha to 128, and LoRA dropout to 0.05.
Regarding the SAM~\cite{kirillov2023segment}, we use EfficientViT-XL1-SAM~\cite{cai2022efficientvit} to accelerate the experiments, and we observe only minor accuracy loss in our pilot study.

\paragraph{Training Details:} Our training is conducted on 8 NVIDIA 32G V100 GPUs, using float16 precision. We employ deepspeed~\cite{aminabadi2022deepspeed} ZeRO2 for multi-GPU distributed training. We use Lion~\cite{chen2024symbolic} as ours optimizer. The learning rate is set to $1\mathrm{e}{-5}$, with a weight decay of $0.1$. We employ a CosineAnnealing learning rate scheduler with a warmup period covering $3\%$ of the total 
steps. The loss function includes only the text cross entropy loss of LLM. The batch size per GPU is $2$, with a gradient accumulation set to $8$. We truncate the maximum text length to $2048$ during training and only train the model for $3$ epochs to prevent overfitting.

\paragraph{Fine-tune SAM light-weight decoder:}
To ensure fair comparison and optimal performance of our model on the COCO extended dataset, we fine-tune our SAM lightweight mask decoder specifically. 
Given the inherent bias in mask annotations within the COCO dataset \cite{kirillov2023segment}, conducting inference directly without fine-tuning SAM's lightweight decoder would not offer a fair comparison against methods that have been trained on COCO with mask decoders. 
Therefore, we fine-tuned our SAM decoder on the COCO dataset for one epoch. Additionally, to prevent data leakage, we excluded from COCO the images present in the RefCOCO, RefCOCO+, RefCOCOg, and gRefCOCO validation and test sets.

\paragraph{PPG Pointing Strategy Detail:}
During the training data generation phase of PPG, we randomly sample $64$ point groups within the ground truth bounding box. Each group consists of two positive points inside the ground truth mask and one negative point outside. We 
keep the $16$ groups with the highest Intersection over Union (IoU) with the ground truth mask, and then randomly pick a single group 
from these. The chosen group of points are encoded into text using the proposed ``mask as prompt'' method to serve as the label for training. During testing, our parser retrieves two positive and one negative points as well.

\paragraph{PQPP Pointing Strategy Detail:}
To train the PQPP, 
we randomly sample $10$ points from the ground truth bounding box. These points are labeled as positive if they fall inside the ground truth mask, and negative otherwise. During testing, we uniformly sample $5 \times 5$ grid of points from the bounding box produced by the MLLM.
We then query the MLLM to determine whether each point lies inside 
the object.
Based on MLLM's response (Yes or No), we filter the outcomes according to the response's confidence level associated with the output token (the probability of emitting that token). We only retain points with a confidence level greater than 0.9 and feed them into SAM to generate the mask.

\subsection{Benchmarks}
\vspace{-5pt}
We use the datasets described in \cref{method:implementatoin_overview} for training. Their test splits are used for evaluation (\textbf{RES dataset}, \textbf{GRES}, \textbf{VQA}). 
In addition, we use \textbf{ReasonSeg}~\cite{lai2023lisa} as a zero-shot evaluation for segmentation from complex reasoning scenarios. This comprehensive evaluation assesses the versatility and effectiveness of our model across a diverse range of referring expression segmentation scenarios.


It is worth mentioning that, compared to other MLLM-based methods, our approach uses significantly less training data. 
A detailed comparison is presented in \cref{tab:training_data_comparison}. For instance, 
GLaMM~\cite{rasheed2023glamm} is trained using the GranD~\cite{rasheed2023glamm} dataset, which has 11M images and 810M object masks. Its annotations are collected through various vision and language models including GPT-4-based rewrites of existing open-source datasets. 
In contrast, our SAM4MLLM only uses a small amount of mask annotation data (100K images, 82K object masks) to enable MLLMs to learn general information, 
but can produce high-quality segmentation masks in conjunction with SAM.

\begin{table}[t]
\centering
\caption{{\bf Comparison of the training data from different methods.} SAM4MLLM uses less training data than other MLLM-based methods, especially in terms of the number of masks.}
\smallcompress
\label{tab:training_data_comparison}
{\resizebox{1\textwidth}{!}{
\begin{tabular}{p{2.2cm}|p{5cm}|p{2cm}|p{3.1cm}}
\toprule
\textbf{Method} & \textbf{Train set w/ mask} & \# img./mask & \textbf{Train set w/o mask} \\
\midrule
LISA\cite{lai2023lisa} & {\scriptsize ADE20K\cite{zhou2019semantic}, COCO-Stuff\cite{caesar2018coco}, PACO-LVIS\cite{ramanathan2023paco}, PartImageNet\cite{he2022partimagenet}, PASCAL-Part\cite{chen2014detect}, refCLEF\cite{rohrbach2016grounding}, refCOCO\cite{yu2016modeling}, refCOCO+\cite{yu2016modeling}, refCOCOg\cite{mao2016generation}} & 150K/1.2M & {\scriptsize LLaVA-Instruct-150k\cite{liu2024visual}} \\
\midrule
PerceptionGPT\cite{pi2023perceptiongpt} & {\scriptsize refCOCO\cite{yu2016modeling}, refCOCO+\cite{yu2016modeling}, refCOCOg\cite{mao2016generation}, Visual Genome\cite{krishna2017visual}, Flicker30k\cite{young2014image}} & 150K/3M & {\scriptsize MSCOCO-Caption\cite{lin2014microsoft}, LLaVA-Instruct-150K\cite{liu2024visual}} \\
\midrule
GSVA\cite{xia2023gsva} & {\scriptsize ADE20K\cite{zhou2019semantic}, COCO-Stuff\cite{caesar2018coco}, PACO-LVIS\cite{ramanathan2023paco}, PartImageNet\cite{he2022partimagenet}, PASCAL-Part\cite{chen2014detect}, refCLEF\cite{rohrbach2016grounding}, refCOCO\cite{yu2016modeling}, refCOCO+\cite{yu2016modeling}, refCOCOg\cite{mao2016generation}, gRefCOCO\cite{liu2023gres}} & 150K/1.2M & {\scriptsize LLaVA-Instruct-150k\cite{liu2024visual}} \\
\midrule
GLaMM\cite{rasheed2023glamm} & {\scriptsize GranD~\cite{rasheed2023glamm} (Automatically annotated for SA-1B) , GranD-f\cite{rasheed2023glamm} (Based on Flickr-30K\cite{young2014image}, RefCOCOg~\cite{mao2016generation}, and PSG\cite{yang2022panoptic})} & 11M/810M & - \\
\midrule
SAM4MLLM (Ours) & {\scriptsize refCOCO\cite{yu2016modeling}, refCOCO+\cite{yu2016modeling}, refCOCOg\cite{mao2016generation}, gRefCOCO\cite{liu2023gres}} & 100K/82K & {\scriptsize VQAv2\cite{antol2015vqa}} \\
\midrule
SAM4MLLM* (Ours) & {\scriptsize refCOCO\cite{yu2016modeling}, refCOCO+\cite{yu2016modeling}, refCOCOg\cite{mao2016generation}, gRefCOCO\cite{liu2023gres}, ADE20K\cite{zhou2019semantic}, 
PACO-LVIS\cite{ramanathan2023paco}, PartImageNet\cite{he2022partimagenet}} & 145K/260K & {\scriptsize VQAv2\cite{antol2015vqa}} \\
\bottomrule
\end{tabular}
}}
\compress
\end{table}

\begin{table}[ht]
\centering
\caption{Comparison of methods on refCOCO, refCOCO+, and refCOCOg datasets.}
\label{tab:results_res}
{\resizebox{\textwidth}{!}{
\begin{tabular}{l ccc ccc cc}
\toprule
\multirow{2}{*}{\textbf{Method}} & \multicolumn{3}{c}{\textbf{refCOCO}} & \multicolumn{3}{c}{\textbf{refCOCO+}} & \multicolumn{2}{c}{\textbf{refCOCOg}} \\ 
\cmidrule(lr){2-4}\cmidrule(lr){5-7}\cmidrule(lr){8-9}
& val & testA & testB & val & testA & testB & val(U) & test(U) \\
\midrule
\emph{\color{gray} LLM based (13B)} \\
{\color{gray}PerceptionGPT-13B [CVPR-24]~\cite{pi2023perceptiongpt}} & {\color{gray} 75.3} & {\color{gray} 79.1} & {\color{gray}72.1} & {\color{gray}68.9} & {\color{gray}74.0} & {\color{gray}61.9} & {\color{gray}70.7} & {\color{gray}71.9} \\
{\color{gray} GSVA-Llama2-13B [CVPR-24]~\cite{xia2023gsva}} & {\color{gray}79.2} & {\color{gray}81.7} & {\color{gray}77.1} & {\color{gray}70.3} & {\color{gray}73.8} & {\color{gray}63.6} & {\color{gray}75.7} & {\color{gray}77.0} \\
\midrule
\emph{traditional methods} \\
MAttNet (CVPR-18)~\cite{yu2018mattnet} & 56.51 & 62.37 & 51.70 & 46.67 & 52.39 & 40.08 & 47.64 & 48.61 \\
STEP [ICCV-19]~\cite{chen2019see} & 60.04 & 63.46 & 57.97 & 48.19 & 52.33 & 40.41 & - & - \\
LSCM [ECCV-20]~\cite{hui2020linguistic} & 61.37 & 64.99 & 59.55 & 49.34 & 53.12 & 43.50 & - & -  \\
VLT [ICCV-21]~\cite{ding2021vision} & 65.65 & 68.29 & 62.73 & 55.50 & 59.20 & 49.36 & 52.99 & 56.65 \\
SeqTR [ECCV-22]~\cite{zhu2022seqtr} & 67.26 & 69.79 & 64.12 & 54.14 & 58.93 & 48.19 & 55.67 & 55.64 \\
CRIS [CVPR-22]~\cite{wang2022cris} & 70.5 & 73.2 & 66.1 & 65.3 & 68.1 & 53.7 & 59.9 & 60.4 \\
LAVT [CVPR-22]~\cite{yang2022lavt} & 72.7 & 75.8 & 68.8 & 62.1 & 68.4 & 55.1 & 61.2 & 62.1 \\
ReLA [CVPR-23]~\cite{liu2023gres} & 73.8 & 76.5 & 70.2 & 66.0 & 71.0 & 57.7 & 65.0 & 66.0 \\
X-Decoder [CVPR-23]~\cite{zou2023generalized} & - & - & - & - & - & - & 64.6 & - \\
PolyFormer-L [CVPR-23]~\cite{liu2023polyformer} & 76.94 & 78.49 & 74.83 & 72.15 & 75.71 & 66.73 & 71.15 & 71.17 \\
VPD [ICCV-2023]~\cite{zhao2023unleashing} & 73.25 & - & - & 62.69 & - & - & 61.96 & - \\
ETRIS [ICCV-2023]~\cite{xu2023bridging} & 71.06 & 74.11 & 66.66 & 62.23 & 68.51 & 52.79 & 60.28 & 60.42 \\
SEEM [NeurIPS-23]~\cite{zou2024segment} & - & - & - & - & - & - & 65.7 & - \\
\midrule
\emph{LLM based (7B)} \\
\color{black}LISA-7B [CVPR-24]~\cite{lai2023lisa} & 74.9 & 79.1 & 72.3 & 65.1 & 70.8 & 58.1 & 67.9 & 70.6 \\
\color{black}PixelLM-7B [CVPR-24]~\cite{ren2023pixellm} & 73.0 & 76.5 & 68.2 & 66.3 & 71.7 & 58.3 & 69.3 & 70.5 \\
\color{black}PerceptionGPT-7B [CVPR-24]~\cite{pi2023perceptiongpt} & 75.1 & 78.6 & 71.7 & 68.5 & 73.9 & 61.3 & 70.3 & 71.7 \\
\color{black}GSVA-7B [CVPR-24]~\cite{xia2023gsva} & {77.2} & 78.9 & {73.5} & 65.9 & 69.6 & 59.8 & 72.7 & 73.3 \\
\color{black}GLaMM-7B [CVPR-24]~\cite{rasheed2023glamm} & {79.5} & \textbf{83.2} & \textbf{76.9} & {72.6} & {\ul 78.7} & {64.6} & {74.2} & {74.9} \\
\midrule
SAM4MLLM-7B-PPG & 76.2 & 80.1 & 72.0 & 71.2 & 75.9 & 64.3 & {74.2} & 74.3 \\
SAM4MLLM-7B-PQPP & 77.1 & {80.9} & 72.5 & {71.5} & {76.8} & {64.7} & {\ul 74.5} & {75.2} \\
\color{black}SAM4MLLM-7B-PQPP-LLaVA1.6 & {\ul 79.6} & {\ul 82.8} & {\ul 76.1} & {\ul 73.5} & {77.8} & {\ul 65.8} & {\ul 74.5} & {\ul 75.6}\\
SAM4MLLM-8B-PQPP-LLaVA1.6 & \textbf{79.8} & 82.7 & 74.7 & \textbf{74.6} & \textbf{80.0} & \textbf{67.2} & \textbf{75.5} & \textbf{76.4} \\
\bottomrule
\end{tabular}
}}
\compress
\compress
\end{table}

\subsection{Main Results}
\vspace{-5pt}
\label{ssec:main_results}
We compare the two variants of our SAM4MLLM, PPG and PQPP, with previous arts on various tasks.
There have been numerous LLM-based methods emerging recently, but our comparisons primarily focus on their results using models of similar scales (7B).

\paragraph{RES dataset:} In \cref{tab:results_res}, we present the performance of PPG and PQPP on the refCOCO datasets~\cite{yu2016modeling}, where our approach outperforms most of the recent LLM-based methods and achieves comparable results to the most recent GLaMM~\cite{rasheed2023glamm}. Additionally, we observe distinct performance variances across datasets. Specifically, our method shows superior results to GLaMM on the RefCOCOg dataset with complex narrative queries, while we obtain inferior results on the RefCOCO and RefCOCO+ dataset with simple short text queries. This advantage likely arises from our model's streamlined architecture, which preserves the language model's comprehension and inference capabilities more effectively, leading to better results on complex queries. 

Notably, our model is trained with nearly $100$ times 
fewer images and $10000$ times 
fewer masks compared to GLaMM, yet it still achieves comparable quality.
The results show that by simply leveraging the connection between off-the-shelf MLLMs and SAM, our model can align MLLMs (already trained with large amounts of multimodal data) to the RES task using considerably less data. Since our model training is consistent with that of the original LLM, our method even performs better than all other methods in the case of longer and more complex sentence understanding (RefCOCOg).

AS for our PPG and PQPP approaches, SAM4MLLM-PQPP outperforms SAM4MLLM-PPG in all cases, as shown in Tab.~\ref{tab:results_res}. The results reveal that 
{\color{black} there is a trade-off between cost and quality when using PPG and PQPP. PPG predicts SAM prompts in once and needs only a single-turn conversation with MLLM. PQPP requires a two-turn conversation for bbox prediction and points classification but achieves better accuracy in most cases. We further use LLaVA1.6 as our MLLM backbone architecture instead of Qwen-VL for PQPP on these datasets. With a more powerful MLLM, the performance can be enhanced.}


\paragraph{GRES:} We present the comparison on the gRefCOCO dataset~\cite{liu2023gres} in \cref{tab:results_gres}. Unlike the RES dataset, this dataset contains the cases where multiple instances or no instances are referred. In this generalized RES task, our method sets the new state-of-the-art among the 7B models on most of the splits and metrics, except for ``Test Set A'', where we lag slightly behind the recent GSVA~\cite{xia2023gsva}.

\paragraph{ReasonSeg:} Our method also demonstrates superior results on the complex reasoning segmentation task, as shown in \cref{table:reasonseg}. It is worth noting that we evaluate on this dataset in a zero-shot manner, meaning our model was not trained on relevant tasks before. {\color{black} Besides, we use more training data to train SAM4MLLM, denoted as SAM4MLLM*. Despite using less training data than LISA, it can outperform LISA-13B-LLaVA1.5.}

\paragraph{VQA:} 
This dataset is not desifned for RES but for visual question answering. We use it to verify that our model, although enhanced by image segmentation functionality, still maintains its original capabilities.
The VQA scores in \cref{tab:results_vqa} demonstrate that our approach does not compromise the VQA abilities acquired during the pre-training phase of our MLLM backbone. In fact, the VQA performance is even boosted, perhaps due to our fine-tuning on more datasets.

\paragraph{PQPP and PPG:}
Our PQPP consistently outperforms PPG on most 
results. We discuss the effect of 
points prompting strategy further in the ablation studies.




\begin{figure}[t]
    \centering
    \includegraphics[width=1\linewidth]{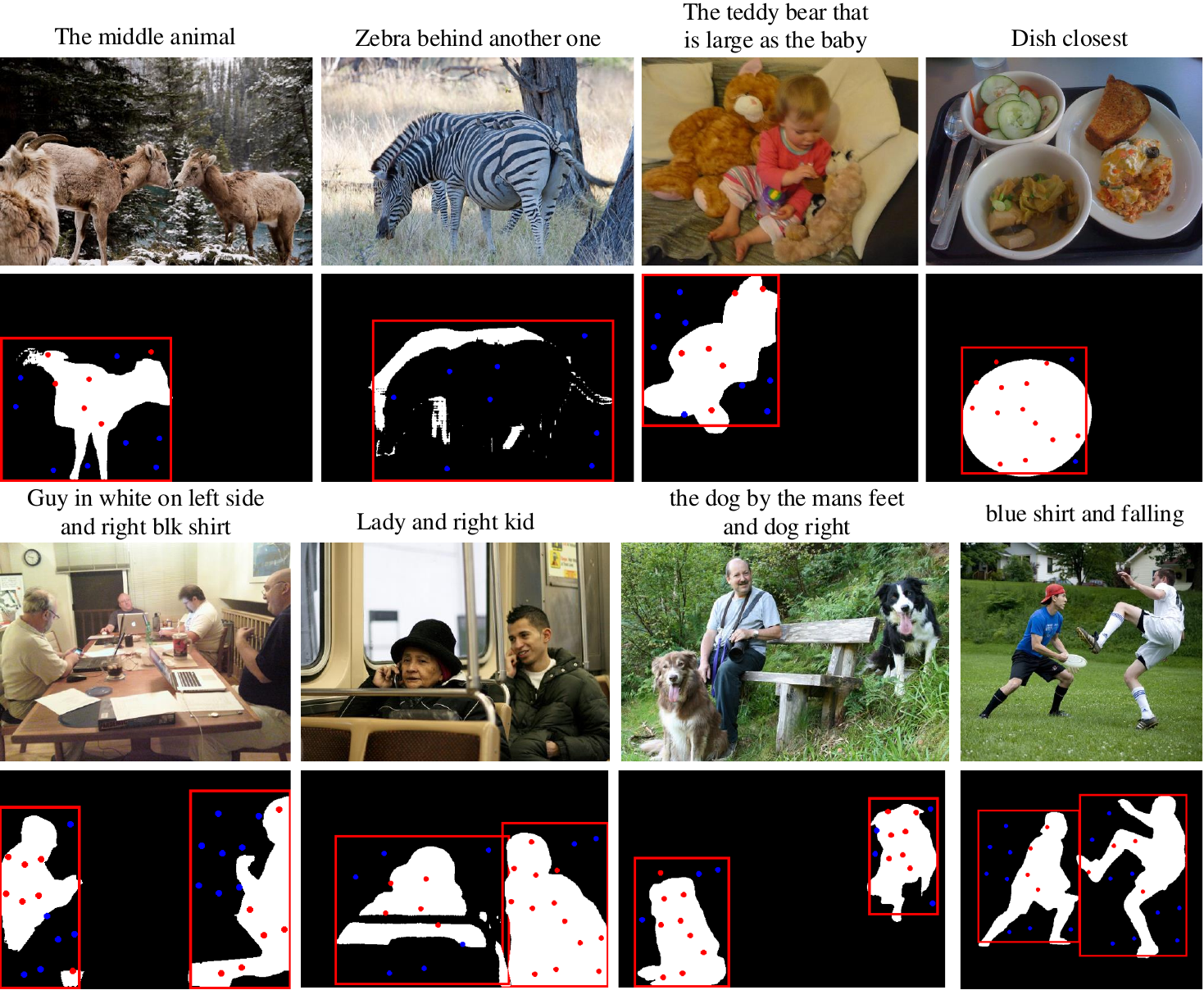}
    \vspace{-1em}
    \caption{Qualitative examples of SAM4MLLM on RES (top) and GRES (bottom) tasks. See \cref{ssec:main_results} for detailed description.}
    \vspace{-1em}
    \label{fig:qualitative_example}
    \compress
\end{figure}

\paragraph{Qualitative results:}
Fig.~\ref{fig:qualitative_example} presents qualitative examples of our SAM4MLLM approach on various referring expression segmentation datasets.
We showcase our results on RES task in the upper row.
The leftest image is from refCOCO, showing the successful segmentation of a specific zebra referred to as ``behind another one.'' The middle-left image, sourced from refCOCO+, demonstrates the accurate identification of the ``middle animal'' among multiple instances. The middle-right image from refCOCOg illustrates the model's ability to handle more complex referring expressions, such as ``The teddy bear that is as large as the baby.'' Finally, the rightest image, also from refCOCO+, showcases the model's understanding of relative positions, correctly segmenting the ``dish closest'' to the referred object.
The bottom row demonstrates SAM4MLLM ability on generalized RES task, where our method accurately segments multiple instances as per the given text.
These examples highlight SAM4MLLM's capability to accurately segment objects based on diverse referring expressions across different datasets.
\vspace{-2.2em}

\begin{table}[t]
\centering
\caption{Results on gRefCOCO(GRES).}
\label{tab:results_gres}
\compress
{\resizebox{\textwidth}{!}{
\begin{tabular}{l ccc @{\hskip 5pt} ccc @{\hskip 5pt} ccc}
\toprule
\multirow{2}{*}{\bf Method} & \multicolumn{3}{c}{\bf Validation Set} & \multicolumn{3}{c}{\bf Test Set A} & \multicolumn{3}{c}{\bf Test Set B} \\
\cmidrule(lr){2-4}\cmidrule(lr){5-7}\cmidrule(lr){8-10}
& gIoU & cIoU & N-acc. & gIoU & cIoU & N-acc. & gIoU & cIoU & N-acc. \\
\midrule
\emph{\color{gray} LLM-based (13B) model} \\
{\color{gray} LISA-13B [CVPR-24]~\cite{lai2023lisa}} & {\color{gray} 65.24} & {\color{gray} 63.96} & {\color{gray}57.49} & {\color{gray} 69.99} & {\color{gray} 71.00} & {\color{gray} 55.43} & {\color{gray} 62.11} & {\color{gray} 62.29} & {\color{gray} 56.34} \\
{\color{gray} GSVA-13B [CVPR-24]~\cite{xia2023gsva}} & {\color{gray} 70.04} & {\color{gray} 66.38} & {\color{gray}66.02} & {\color{gray}73.29} & {\color{gray}72.79} & {\color{gray}64.72} &{\color{gray} 65.45} & {\color{gray} 63.20} & {\color{gray} 62.47} \\
\midrule
\emph{traditional methods} \\
MAttNet [CVPR-18]~\cite{yu2018mattnet} & 48.24 & 47.51 & 41.15 & 59.30 & 58.66 & 44.04 & 46.14 & 45.33 & 41.32 \\
LTS [CVPR-21]~\cite{jing2021locate} & 52.70 & 52.30 & - & 62.64 & 61.87 & - & 50.42 & 49.96 & - \\
VLT [ICCV-21]~\cite{ding2021vision} & 52.00 & 52.51 & 47.17 & 63.20 & 62.19 & 48.74 & 50.88 & 50.52 & 47.82 \\
CRIS [CVPR-22]~\cite{wang2022cris} & 56.27 & 55.34 & - & 63.42 & 63.82 & - & 51.79 & 51.04 & - \\
LAVT [CVPR-22]~\cite{yang2022lavt} & 58.40 & 57.64 & 49.32 & 65.90 & 65.32 & 49.25 & 55.83 & 55.04 & 48.46 \\
ReLA [CVPR-23]~\cite{liu2023gres} & 63.60 & 62.42 & 56.37 & 70.03 & 69.26 & 59.02 & 61.02 & 59.88 & 58.40 \\
\midrule
\emph{LLM based (7B)} \\
\color{black}LISA-7B [CVPR-24]~\cite{lai2023lisa} & 61.63 & 61.76 & 54.67 & 66.27 & 68.50 & 50.01 & 58.84 & 60.63 & 51.91 \\
\color{black}GSVA-7B [CVPR-24]~\cite{xia2023gsva} & 66.47 & 63.29 & 62.43 & {\ul 71.08} & {69.93} & 65.31 & 62.23 & 60.47 & 60.56 \\
\midrule
SAM4MLLM-7B-PPG & {68.37} & {65.66} & {\ul63.71} & 69.05 & 69.62 & \textbf{65.96} & {63.71} & {62.35} & {\ul61.25} \\
SAM4MLLM-7B-PQPP & {\ul68.96} & {\ul 66.33} & {62.96} & {70.54} & {\ul70.13} & {\ul65.82} & {\ul 63.98} & {\ul63.21} & \textbf{61.61} \\
SAM4MLLM-8B-PQPP-LLaVA1.6 & \textbf{71.86} & \textbf{67.83} & \textbf{66.08} & \textbf{74.15} & \textbf{72.22}& 63.92 & \textbf{65.29} & \textbf{63.42} & 59.99 \\
\bottomrule
\end{tabular}
}}
\compress
\end{table}


\begin{table}[t]
\vspace{-6pt}
\begin{minipage}[t]{0.55\linewidth}
\centering
\caption{Results on ReasonSeg(Zero-Shot).}
\label{tab:results_reasonseg}
\begin{tabular}{lcc}
\toprule
\multirow{2}{*}{\textbf{Method}} & \multicolumn{2}{c}{\textbf{val}} \\
\cmidrule{2-3}
& gIoU & cIoU \\
\midrule
{\color{gray} LISA-13B [CVPR-24]~\cite{lai2023lisa}} & {\color{gray} 48.9} & {\color{gray} 46.9} \\
{\color{gray} LISA-13B-LLaVA1.5 [CVPR-24]~\cite{lai2023lisa}} & {\color{gray} 57.7} & {\color{gray} 60.3} \\
OVSeg [CVPR-23] \cite{liang2023open} & 28.5 & 18.6 \\
GRES [CVPR-23] \cite{liu2023gres} & 22.4 & 19.9 \\
X-Decoder [CVPR-23] \cite{zou2023generalized} & 22.6 & 17.9 \\
SEEM [NeurIPS-23] \cite{zou2024segment} & 25.5 & 21.2 \\
\color{black} LISA-7B [CVPR-24]~\cite{lai2023lisa} & 44.4 & 46.0 \\
\color{black} LISA-7B-LLaVA1.5 [CVPR-24]~\cite{lai2023lisa} & 53.6 & 52.3 \\
\midrule
SAM4MLLM-7B-PPG & {46.1} & {47.9} \\
SAM4MLLM-7B-PQPP & \underline{46.7} & \underline{48.1} \\
SAM4MLLM*-8B-PQPP-LLaVA1.6 & \textbf{58.4} & \textbf{60.4} \\
\bottomrule
\end{tabular}
\label{table:reasonseg}
\end{minipage}
\hfill
\begin{minipage}[t]{0.4\linewidth}
\centering
\caption{Result on Vision Question Answering}
\label{tab:results_vqa}
\begin{tabular}{lc}
\toprule
\textbf{Method} & \textbf{VQAv2} \\
\midrule
Qwen-VL-Chat-7B \cite{bai2023qwen} & 78.2 \\
SAM4MLLM-7B-PPG  & {\bf 78.7} \\
SAM4MLLM-7B-PQPP & {\bf 78.7} \\
\bottomrule
\end{tabular}
\label{table:abl_point_sampling}
\end{minipage}
\end{table}

\begin{table}[t]
\centering
\caption{Ablation studies of SAM4MLLM-PQPP.}
\vspace{-1.5em}
\begin{subtable}[t]{0.44\linewidth}
\centering
\caption{Effect of point filter threshold.}
\vspace{-.8em}
\label{tab:ablation_point_threshold}
\begin{tabular}{lc}
\toprule
\textbf{Method} & \textbf{RefCOCOg val} \\
\midrule
0.6 & 69.5 \\
0.7 & 70.8 \\
0.8 & 72.2 \\
0.9 & \textbf{74.5} \\
0.95 & 73.8 \\
\bottomrule
\end{tabular}
\label{table:abl_threshold}
\end{subtable}
\hfill
\begin{subtable}[t]{0.55\linewidth}
\centering
\caption{Effect of point sampling strategy in inference.}
\vspace{-.8em}
\label{tab:ablation_point_sampling}
\begin{tabular}{lc}
\toprule
\textbf{Method} & \textbf{RefCOCOg val} \\
\midrule
5x5 grid (N=25) & \textbf{74.5} \\
6x6 grid (N=36) & 74.4 \\
random sample (N=25) & 73.7 \\
random sample (N=36) & 74.2 \\
\bottomrule
\end{tabular}
\label{table:abl_point_sampling}
\end{subtable}
\compress
\end{table}

\section{Ablation Study}
\vspace{-6pt}

To gain a deeper understanding of the factors contribution, we conducted some ablation studies focusing on the best-performing variant, SAM4MLLM-PQPP. Our investigations centered around the following aspects: confidence threshold for filtering points in PQPP and sampling strategy for selecting points within the bounding box, providing insights into our method's robustness and adaptability.

\paragraph{Points Filtering Threshold}
First, we examined the impact of the confidence threshold used in PQPP to filter points based on the MLLM's responses. We experimented with threshold values ranging from 0.6 to 0.95 on the RefCOCOg validation set and evaluated their effect on the cIoU metric. Our results, presented in Table \ref{tab:ablation_point_threshold}, reveal that a threshold of 0.9 strikes the optimal balance, as further increasing or decreasing the threshold leads to a notable decline in cIoU. This finding highlights the importance of carefully tuning the confidence threshold to ensure the best possible segmentation quality.

\paragraph{Points Sampling Strategy}
Next, we explored the influence of the point sampling strategy within the bounding box on the overall performance. We compared two approaches: grid-based sampling and random sampling, while also varying the number of sampled points. As shown in Table \ref{tab:ablation_point_sampling}, a 5x5 grid sampling pattern consistently yields the highest accuracy. The result suggests that a uniform distribution of points across the bounding box provides the most informative cues for the MLLM to accurately determine the object's location and shape.

\section{Conclusion}
\vspace{-8pt}

In this paper, we introduced SAM4MLLM, a novel approach that integrates the Segment Anything Model (SAM) with Multi-Modal Large Language Models (MLLMs) to address the Referring Expression Segmentation (RES) task. By encoding object masks as discrete text prompts, our method enables MLLMs to understand and generate pixel-level object localization information without requiring complex architectural modifications or additional loss functions. Our method is simple but effective. Through experiments on various RES benchmarks, we demonstrate that SAM4MLLM achieves competitive performance while maintaining the simplicity and generalizability of the original language models. 
Our work explores a new direction for leveraging the capabilities of foundation models to tackle complex vision-language tasks in a more streamlined and unified manner. We hope that the insights gained from this research will inspire further investigations into effectively combining the strengths of different models to solve challenging multimodal problems. Future work could involve extending our approach to handle a broader range of visual reasoning tasks and conducting more in-depth analyses to better understand the interplay between language models and visual foundation models.


%
%
%

\bibliographystyle{splncs04}
\bibliography{main}
\title{Supplementary Materials\\ SAM4MLLM: Enhance Multi-Modal Large Language Model for Referring Expression Segmentation} 

\titlerunning{SAM4MLLM}

\author{Yi-Chia Chen\inst{1} \and Wei-Hua Li\inst{1} \and 
Cheng Sun\inst{2} \and \\ Yu-Chiang Frank Wang \inst{1,2} \and Chu-Song Chen\inst{1}}

\authorrunning{Chen et al.}

\institute{National Taiwan University, Taipei, Taiwan. \\
\email{\{r11922a04, d12922009, chusong\}@csie.ntu.edu.tw}\\ \and
NVIDIA, Taipei, Taiwan. \\
\email{\{chengs, frankwang\}@nvidia.com}
}

\maketitle

\section{\color{black}Details of point sampling strategies.}
{In Tab.~\ref{tab:upper_bound}, we provide an analysis for using SAM as our backend, where the upper-bound is the maximum IoU from multiple SAM prompts sampled using the ground-truth masks.
The upper-bound is around 87.8\% IoU, which is much higher than all existing methods.
The predicted SAM prompts by our method achieve around 75\% IoU, suggesting there is room for improvement on the MLLM side toward reaching the upper-bound quality.
In Tab.~\ref{tab:different_k}, we provide additional ablations of PPG, where more groups {\footnotesize ($k$)} or more points {\footnotesize ($n_1,n_2$)} result in worse accuracy.
The results support our claim in the main paper that it is challenging for MLLM to learn to predict multiple points simultaneously.}

\begin{table}
    \centering
        \begin{tabular}{@{}c|ccc@{}}
            \hline
             &RefCOCOg val & RefCOCOg test \\ \hline
            Upper Bound &  87.9    &    87.8   \\ \hline
            SAM4MLLM &   74.5     &    75.2   \\ \hline
    \end{tabular}
    \vspace{1em}
    \caption{Comparison between upper bound and SAM4MLLM.}
    \label{tab:upper_bound}
\end{table}

\begin{table}
    \centering
    \begin{tabular}{@{}c|ccccc@{}}
            \hline
            $(k,n_1,n_2)$ & (1,2,1) & (2,2,1) & (3,2,1) & (1,3,2) & (1,4,4) \\
            \hline
            RefCOCOg val & \textbf{72.0} & 71.6 & 71.7 & {\ul 71.9} & 71.4 \\
            \hline
            RefCOCOg test &\textbf{72.9} & 72.6 & 72.2 & {\ul 72.7} & 72.2 \\
            \hline
    \end{tabular}
    \vspace{1em}
    \caption{Ablation on different k, n1 and n2.}
    \label{tab:different_k}
\end{table}

\section{\color{black}Details of SAM in SAM4MLLM}
\begin{figure}[t]
    \centering
    \includegraphics[width=1\linewidth]{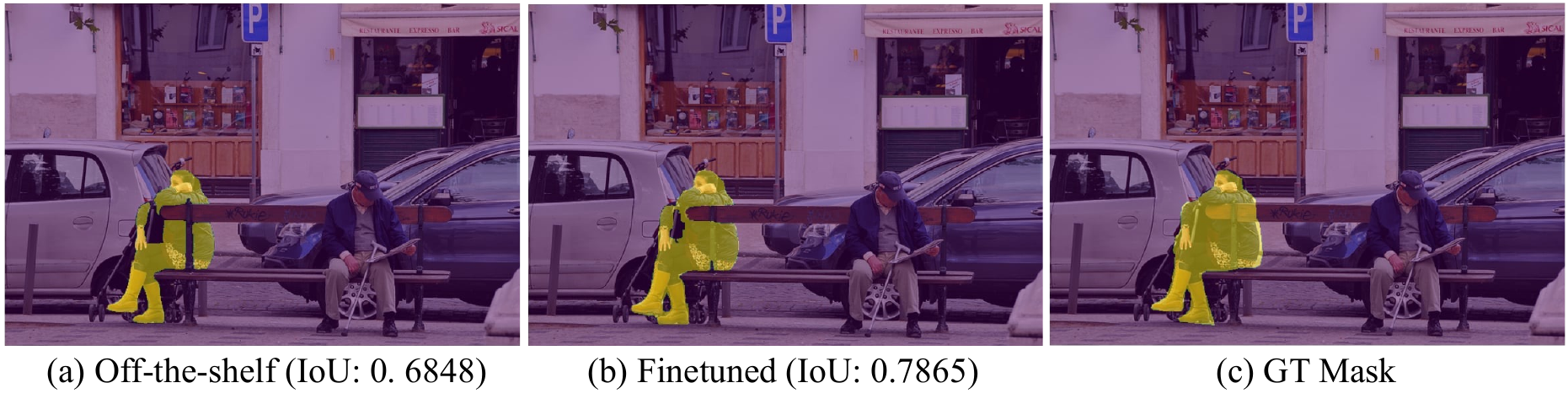}
    \caption{(a) off-the-shelf SAM (b) finetuned SAM (c) GT}
    \label{fig:finetune_SAM}
\end{figure}

Following LISA~\cite{lai2023lisa} and GSVA~\cite{xia2023gsva}, we finetune SAM's decoder on COCO extended datasets (RefCOCO and GRES), while using off-the-shelf SAM on ReasonSeg. This is due to the inherent bias in mask annotations within the COCO dataset.
As shown in Tab.~\ref{tab:SAM_finetune} and Fig.~\ref{fig:finetune_SAM}, with finetuned SAM, SAM4MLLM can predict more consistent segmentation mask that matches the granularity of COCO where coarser masks are often provided as ground truths.

\begin{table}
    \centering
        \begin{tabular}{@{}c|cc@{}}
            \hline
            SAM          &    RefCOCOg val      & RefCOCOg test \\ \hline
            w/o finetene &     72.9     &      73.3     \\ \hline
            w/ finetune  &     74.5     &      75.2     \\ \hline
    \end{tabular}
    \vspace{1em}
    \caption{Comparison between off-the-shelf and finetuned SAM.}
    \label{tab:SAM_finetune}
\end{table}

\section{\color{black}Ablation regarding the dataset combination.}
If we only use RES datasets for training without using the non-segmentation VQA datase, the segmentation performance is not influenced as shown in Tab.~\ref{tab:different_training_data}. Moreover, our method can preserve VQA capabilities when using both datasets.

\begin{table}
    \centering
    {
    \begin{tabular}{@{}c|ccc@{}}
    \hline
             &RefCOCOg val & RefCOCOg test & VQA \\ \hline
            RES &    72.9    &    73.6    &  0.0 \\ \hline
            RES + VQA &   73.0   &  73.6  &  78.6  \\ \hline
    \end{tabular}
    }
    \vspace{1em}
    \caption{Comparison of the different training data.}
    \label{tab:different_training_data}
\end{table}

\section{Segmentation Samples}
\begin{figure}[!ht]
    \centering
    \includegraphics[width=1\linewidth]{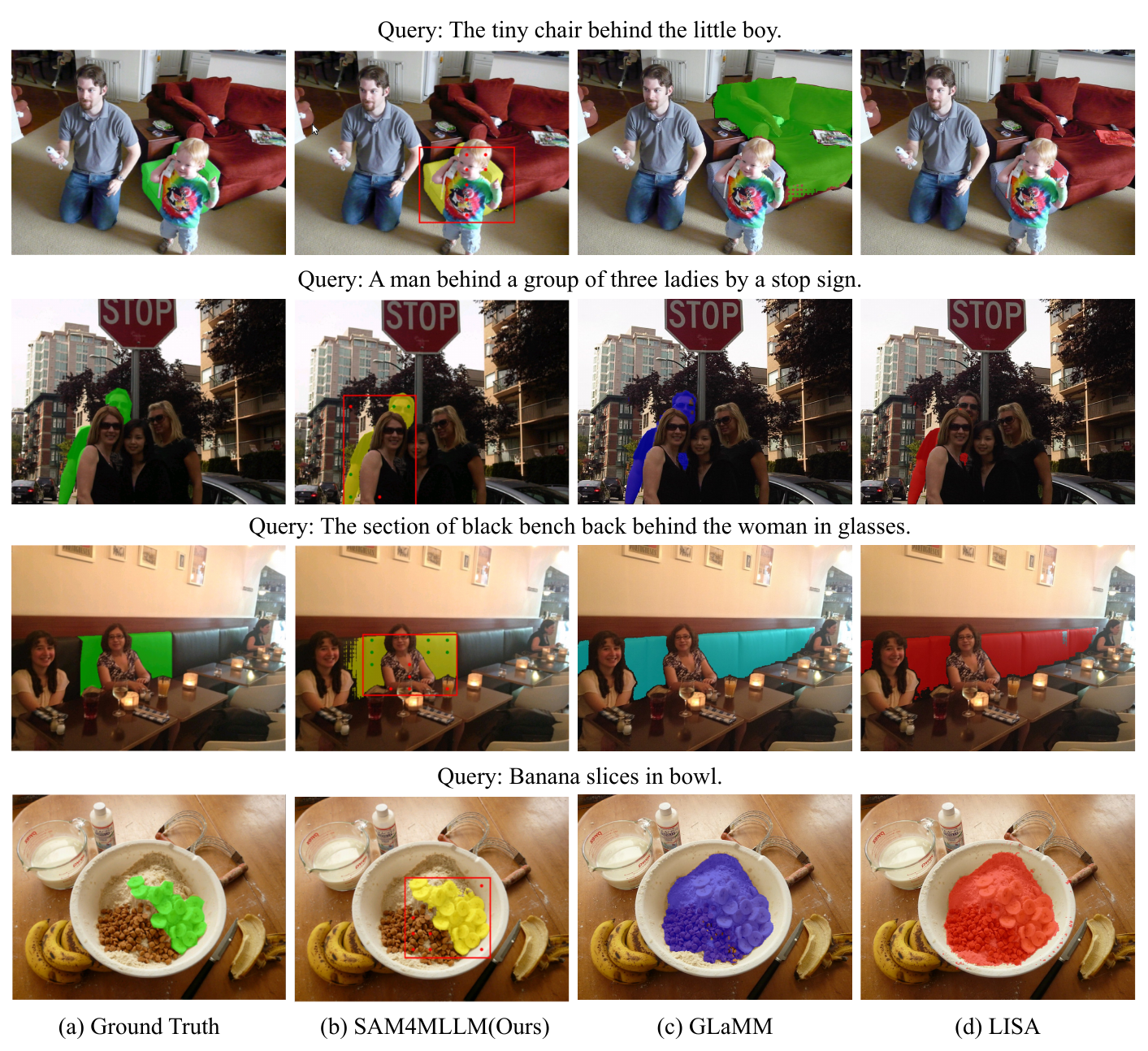}
    \caption{Visaul comparisons between (a) Ground Truth, (b) SAM4MLLM(ours), (c) GLaMM~\cite{rasheed2023glamm} and (d) LISA~\cite{lai2023lisa} on RefCOCOg.}
    \label{fig:RefCoCog}
\end{figure}

We present a visual comparison of SAM4MLLM with GLaMM~\cite{rasheed2023glamm} and LISA~\cite{lai2023lisa} on the RefCOCOg~\cite{mao2016generation} dataset, which contains longer queries than RefCOCO~\cite{yu2016modeling} and RefCOCO+~\cite{yu2016modeling}, and is more challenging in referring expression segmentation. 
As we can see in Fig.\ref{fig:RefCoCog}, our approach better captures the intent of the query and predicts more accurate segments. In contrast, GLaMM and LISA may segment incomplete or larger objects instead of the regions specified in the query.

\begin{figure}[!ht]
    \centering
    \includegraphics[width=0.9\linewidth]{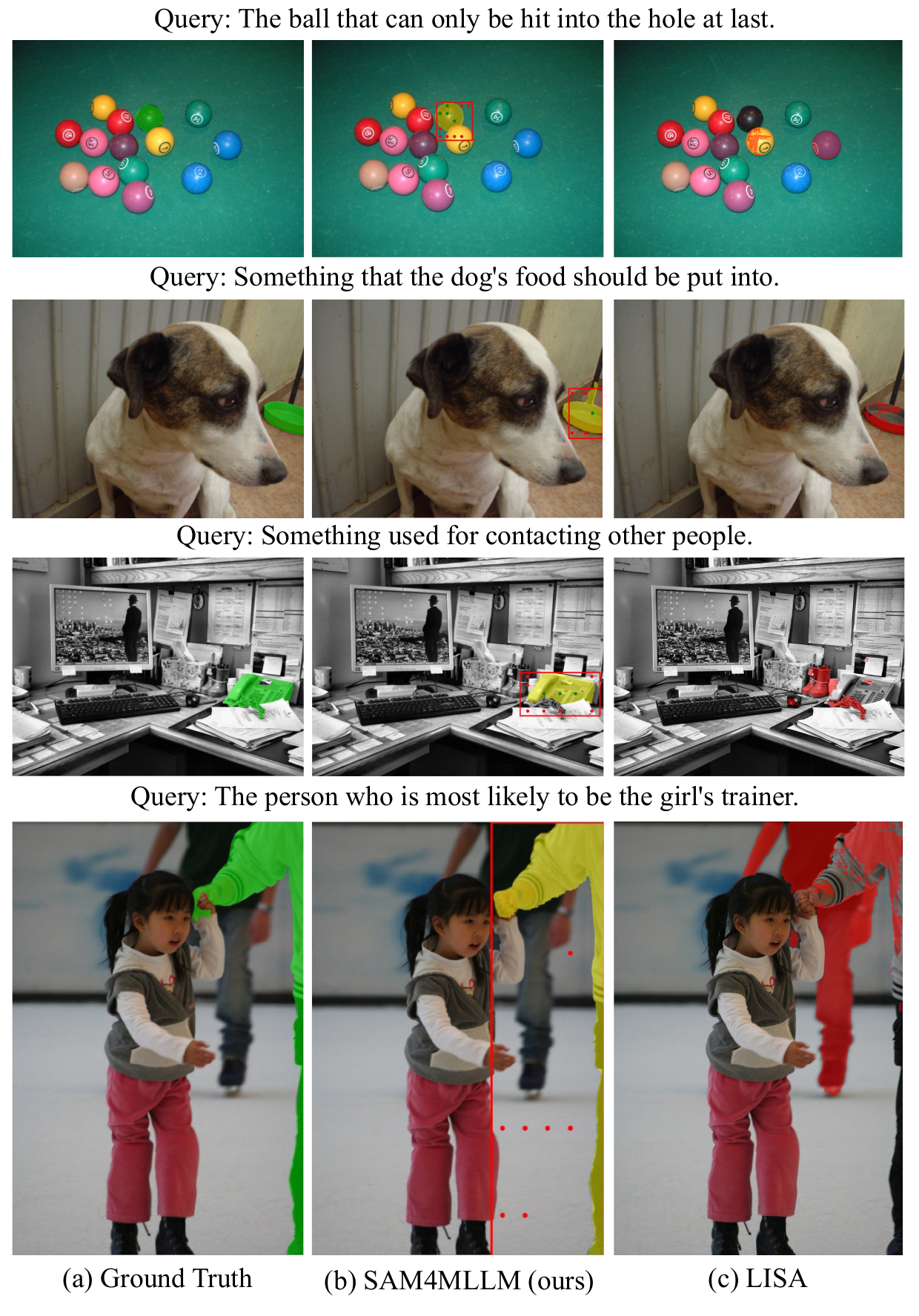}
    \caption{Visual comparisons between (a) Ground Truth, (b) SAM4MLLM(ours) 
    and (c) LISA~\cite{lai2023lisa} on ReasonSeg.}
    \label{fig:ReasonSeg}
\end{figure}

Fig.~\ref{fig:ReasonSeg} illustrates the comparison between SAM4MLLM and LISA on ReasonSeg, a dataset that includes implicit query descriptions without providing specific object names, thus testing the models' inference abilities. Despite this challenge, our SAM4MLLM still demonstrates robust segmentation capability.

\section{Discussion}
\paragraph{Limitation:} 
A limitation of our approach is that it depends on SAM for the final segmentation output, thus the quality of the generated masks is inherently constrained by SAM's segmentation capability. However, our approach will also improve effortlessly with the progress of SAM.




\paragraph{Future work:} Integrating SAM4MLLM with other state-of-the-art segmentation models or developing techniques to refine SAM's outputs in challenging scenarios are worth exploring. 
SAM4MLLM offers significant advantages by combining the strengths of MLLMs and SAM in a straightforward yet effective manner. The MLLM's ability to understand and reason about referring expressions, combined with SAM's strong segmentation capabilities, enables SAM4MLLM to tackle complex referring expression segmentation tasks. As research in both language models and segmentation models continues to advance, we expect the performance of SAM4MLLM to be further improved.

\section{Image License}
In this paper, we use several images from RefCOCO, RefCOCOg, and RefCOCO+ for illustration. The images from RefCOCO, RefCOCOg, and RefCOCO+ are sourced from the COCO dataset. We organized its sources and flickr URLs in Table~\ref{tab:license}.

\begin{table*}
    \centering
        \begin{tabular}{|c|l|c|c|c|}
\hline
\multicolumn{2}{|c|}{Figure}     & \multicolumn{2}{c|}{Source} & flickr URL \\ \hline
\multirow{10}{*}{Main Paper}   & Fig.1\&2(a) &\multicolumn{1}{c|}{\multirow{14}{*}{\begin{tabular}[c]{@{}c@{}}COCO\\ Train set\end{tabular}}}&  580837.jpg      & \href{http://farm6.staticflickr.com/5050/5225690150_731eccfdc4_z.jpg}{Link}     \\ \cline{2-2} \cline{4-5} 
                               & Fig.1\&2(b) &\multicolumn{1}{c|}{}&  084259.jpg      &\href{http://farm6.staticflickr.com/5011/5427608106_73cec60aaf_z.jpg}{Link}  \\ \cline{2-2} \cline{4-5} 
                               & Fig.3(leftest-top) &\multicolumn{1}{c|}{}& 161757.jpg&\href{http://farm9.staticflickr.com/8560/8700300305_cef976c4f9_z.jpg}{Link}   \\ \cline{2-2} \cline{4-5} 
                               & Fig.3(middle-left-top) &\multicolumn{1}{c|}{}& 073387.jpg &
                               \href{http://farm5.staticflickr.com/4108/4845987782_89f01438d4_z.jpg}{Link} \\ \cline{2-2} \cline{4-5} 
                               & Fig.3(middle-right-top) &\multicolumn{1}{c|}{}&  066669.jpg  &
                               \href{http://farm4.staticflickr.com/3325/3274392233_a9d55f2400_z.jpg}{Link}   \\ \cline{2-2} \cline{4-5} 
                               & Fig.3(rightest-top) &\multicolumn{1}{c|}{}&  084259.jpg      &
                               \href{http://farm6.staticflickr.com/5011/5427608106_73cec60aaf_z.jpg}{Link} \\ \cline{2-2} \cline{4-5} 
                               & Fig.3(leftest-bottom) &\multicolumn{1}{c|}{}& 001947.jpg    &\href{http://farm4.staticflickr.com/3513/3935531621_a1628aea1f_z.jpg}{Link}   \\ \cline{2-2} \cline{4-5} 
                               & Fig.3(middle-left-bottom) &\multicolumn{1}{c|}{}& 008657.jpg  &\href{http://farm3.staticflickr.com/2765/4281004040_0a880b95a9_z.jpg}{Link}   \\ \cline{2-2} \cline{4-5} 
                               & Fig.3(middle-right-bottom) &\multicolumn{1}{c|}{}&  002400.jpg      &\href{http://farm5.staticflickr.com/4111/5087340385_55b12cbf58_z.jpg}{Link}  \\ \cline{2-2} \cline{4-5} 
                               & Fig.3(rightest-bottom) &\multicolumn{1}{c|}{}&  000839.jpg      &\href{http://farm3.staticflickr.com/2422/3626647125_aaf6626d7e_z.jpg}{Link}  \\ \cline{1-2} \cline{4-5}
\multirow{5}{*}{Supplementary} & Fig.1    &\multicolumn{1}{c|}{}&  580668.jpg   &\href{http://farm6.staticflickr.com/5323/9626235495_344ded3915_z.jpg}{Link}  \\ \cline{2-2} \cline{4-5}
                               & Fig.2 (top)  &\multicolumn{1}{c|}{}&  141101.jpg  &\href{http://farm3.staticflickr.com/2487/3793910440_ac06c2b21d_z.jpg}{Link}  \\ \cline{2-2} \cline{4-5}
                               & Fig.2 (middle top)  &\multicolumn{1}{c|}{}& 058633.jpg  &\href{http://farm4.staticflickr.com/3398/4601942847_f3f70d0854_z.jpg}{Link}  \\ \cline{2-2} \cline{4-5}
                               & Fig.2 (middle bottom)  &\multicolumn{1}{c|}{}& 220037.jpg   &\href{https://farm4.staticflickr.com/3677/9091983943_8260b4065a_z.jpg}{Link}   \\ \cline{2-2} \cline{4-5}
                               & Fig.2 (bottom)  &\multicolumn{1}{c|}{}& 465457.jpg   &\href{http://farm8.staticflickr.com/7081/7081669181_ea2b168664_z.jpg}{Link}  \\ \hline 
\end{tabular}
    \vspace{1em}
    \caption{Image Source and flickr URL}
    \label{tab:license}
\end{table*}

\end{document}